\documentclass[12pt]{article}
\usepackage{amsmath}
\usepackage{times}
\usepackage{graphicx}
\usepackage{color}
\usepackage{multirow}
\usepackage[authoryear]{natbib}
\usepackage{rotating}
\usepackage{bbm}
\usepackage{latexsym}
\newcommand{\captionfonts}{\normalsize}

\makeatletter  
\long\def\@makecaption#1#2{%
  \vskip\abovecaptionskip
  \sbox\@tempboxa{{\captionfonts #1: #2}}%
  \ifdim \wd\@tempboxa >\hsize
    {\captionfonts #1: #2\par}
  \else
    \hbox to\hsize{\hfil\box\@tempboxa\hfil}%
  \fi
  \vskip\belowcaptionskip}
\makeatother   
\usepackage{graphicx}
\usepackage[dvipsnames]{xcolor}

\usepackage{graphicx}
\usepackage{float}
\usepackage{amsmath}
\usepackage{amssymb}
\usepackage[printonlyused]{acronym}
\usepackage{mathtools}
\usepackage[bottom]{footmisc}
\usepackage{xfrac}
\usepackage[dvipsnames]{xcolor}
\graphicspath{{./figs/}}
\usepackage{natbib}

\begin{document}
\newacro{pam}[PaM]{Prime and Modulate}
\newacro{GDM}[GDM]{Gradient Descent Method}
\newacro{ldr}[LDR]{Light Dependent Resistor}
\newacro{gsv}[GSV]{Grey-Scale Value}
\newacro{RL}[RL]{Reinforcement Learning}
\newacro{DL}[DL]{Deep Learning}
\newacro{DNN}[DNN]{Deep Neural Network}
\newacro{BP}[BP]{back-propagation}
\newacro{RPi}[RPi]{Raspberry Pi 3B+}
\newacro{EVGP}[EVGP]{exploding and vanishing gradient problem}
\newacro{LSTM}[LSTM]{long short-term memory}

\hspace{13.9cm}1

\ \vspace{20mm}\\

{\LARGE Prime and Modulate Learning: Generation of forward models 
with signed back-propagation and environmental cues}

\ \\
{\bf \large Sama 
Daryanavard$^{\displaystyle 1},$  Bernd Porr$^{\displaystyle 1}$}\\
{$^{\displaystyle 1}$Biomedical Engineering Division, School of 
Engineering, University of Glasgow, Glasgow G12 8QQ, UK.}\\

%\ \\[-2mm]
%{\bf Keywords:} 

\thispagestyle{empty}
\markboth{}{NC instructions}
\ \vspace{-0mm}\\
%
%Abstract
\begin{center} {\bf Abstract} \end{center}
Deep neural networks employing error back-propagation for learning can
suffer from exploding and vanishing gradient problems. Numerous
solutions have been proposed such as normalisation techniques or
limiting activation functions to linear rectifying units. In this work
we follow a different approach which is particularly applicable to
closed-loop learning of forward models where back-propagation makes
exclusive use of the \textsl{sign} of the error signal to prime the
learning, whilst a global relevance signal modulates the rate of
learning. This is inspired by the interaction between local plasticity
and a global neuromodulation. For example, whilst driving on an empty
road, one can allow for slow step-wise optimisation of actions,
whereas, at a busy junction, an error must be corrected at
once. Hence, the error is the priming signal and the intensity of the
experience is a modulating factor in the weight change. The advantages
of this Prime and Modulate paradigm is twofold: it is free from
normalisation and it makes use of relevant cues from the
environment to enrich the learning. We present a mathematical
derivation of the learning rule in z-space and demonstrate the
real-time performance with a robotic platform. The results show a
significant improvement in the speed of convergence compared to that
of the conventional back-propagation.
%%%%%%%%%%%

\section{Introduction}

Since its inception, deep learning has proven remarkably successful in
a wide variety of areas, such as: image classification, speech
recognition, and reinforcement learning
\cite{bahri2020statistical}. \acfp{DNN} employ activation functions,
such as the logistic or the hyperbolic tangent ($\tanh$), and are most
commonly trained using an error signal with the \ac{GDM} for
optimisation. The derivative of such functions have a narrow range of
practical values and a limit of zero outwith, therefore the
propagation of the error signal through this non-linearity often
suffers from the \acf{EVGP}. This has ignited a significant research
effort into addressing this issue; amongst the solutions are the
network architectures such as \acf{LSTM}, precise weight
initialisations, and specific non-linear activation functions
\citep{hanin2018neural}. Notably, linear rectifying units are used to
remedy this problem which effectively remove the derivative and thus,
drastically alter the behaviour of the network and the nature of
learning. On the other hand, the logistic function allows for a smooth
saturation of signals, along with $\tanh$. This is in accordance with
neuroscience where nearly all psychometric functions or internal
neuronal processing follow a sigmoid activation pattern.

From a neurophysiological standpoint, learning is driven by local and
global mechanisms changing synaptic plasticity \cite{Reynolds2002}.
In particular, in closed-loop learning an interplay of local learning
and global learning which has been advantageous, for example,
improving the stability of learning when generating a forward model 
of a
reflex \cite{Porr2007}. Nonetheless, to date, this class of
closed-loop learning has only been used in shallow networks, often 
with a single learning units or shallow networks
\citep{Porr2007,kulvicius2007chained,maffei2017perceptual}.

In this work, we present a learning paradigm that combines local error
back-propagation and global modulation to create a robust learning
scheme for the generation of forward models. More specifically, only
the sign of the propagating signal is used to prime the nature of
weight change in the context of their local connections, whilst a
global ``relevance'' signal acts as a third factor to excite the
weight changes across the network. This offers not only a robust,
nearly one-shot real-time learning, free of \acf{EVGP}, but also a
more comparable learning model to neuroscience than the conventional
\acl{BP}. In this paper, we implement this novel algorithm on a
physical robot that learns to improve the performance of a closed-loop
feedback controller (a \textit{reflex}) by calculating its forward
model, and thus prevents the triggering of the controller.

\section{The closed-loop learning platform}
\begin{figure}[H]
  \centering
  \includegraphics[width=1\textwidth]{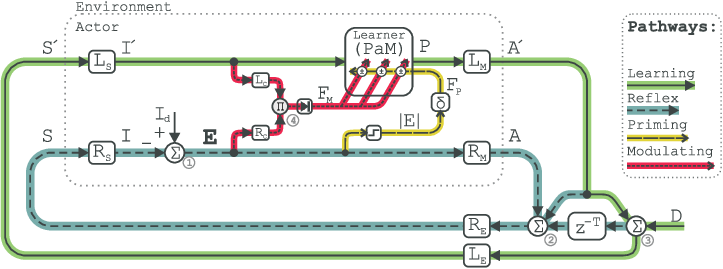}  
  \caption{\textit{The \acf{pam} closed-loop learning platform 
  consisting of an inner reflex loop (highlighted in blue) and an 
  outer learning loop (highlighted in green). The signal and transfer 
  function notations are as follow: Disturbance (D), delay function 
  ($z^{-T}$), reflex environment ($R_{E}$), state of reflex ($S$), 
  reflex sensory unit ($R_{S}$), reflex sensory input ($I$), desired 
  sensory input ($I_{d}$), error signal ($ E$), reflex motor unit 
  ($R_{M}$), reflex motor action ($A$), learner environment 
  ($L_{E}$), state of learner ($S^{\prime}$), learner sensory unit 
  ($L_{S}$), leaner sensory input ($I^{\prime}$), learner ($\ac{pam}$ 
  network), learner motor action ($A^{\prime}$). The priming pathway 
  is highlighted with red where $F_{P}$ is the priming factor that 
  propagates through the network. Modulating pathway is highlighted 
  with yellow, where, reflex cue function ($R_{C}$) and learner cue 
  function ($L_{C}$) lead to the modulating factor ($ F_{M}$) and 
  drives the learning.  }}\label{fig:closedloop}
\end{figure}

\paragraph{The reflex and predictive loops:}
Figure~\ref{fig:closedloop} shows the learning platform.
The so called ``reflex'' is a reactive closed loop controller which
aims to stay as close to its desired state $I_d$ as possible. This
fixed reflex controller acts against the disturbance $D$ which travels
through the environmental transfer function $R_E$. This leads to a new
 state $S$ that is picked up by $R_{S}$ and causes a sensor
signal $I$. This signal is compared to the desired state $I_d$ at node
\textcircled{\small{1}} and in
turn creates the error signal $E$. This error signal is translated
into an appropriate action $A$ via the motor transfer function $R_M$ 
to
counteract the disturbance $D$ at node
\textcircled{\small{2}}.
Here, the crucial aspect of the error signal $E$
is that it can be used for
\textsl{realtime} learning that enables the learner
to generate a forward model of the reflex loop.

Learning of the forward model is performed by the novel \ac{pam} 
network placed in the learning (predictive) loop. The disturbance 
travels through the learner's environment $L_{E}$, leading to a new 
state $S^\prime$ which is picked up by learner's sensory unit $L_{S}$ 
and causes a predictive sensory
signal $I^\prime$. This is fed into the learning algorithm which in 
turn generates the output $P$. This is translated into the predictive 
action $A^\prime$, executed by the learner's motor unit $L_{M}$, to 
eliminate the Disturbance $D$ at node \textcircled{\small{3}}, before 
it can enter the reflex loop.

\paragraph{z-space:} The signals in Figure~\ref{fig:closedloop} are 
discrete time and real-valued physical measurements, therefore are 
more accurately referred to as sequences. Due to the recursive nature 
of closed-loop systems it is beneficial to analyse their behaviour in 
z-space where discrete time-domain sequences are transformed into 
complex frequency-domain representations 
\citep{oppenheim1999discrete}. We use the unilateral z-transform 
representation of these sequences in this work, which, for an 
arbitrary sequence $x[n]$, is defined as:

\begin{align}
	X(z) = \mathcal{Z}\{x[n]\} =  \Sigma_{n=0}^{\infty}(x[n]z^{-n})
\end{align}

Where $\mathcal{Z}\{.\}$ is z-transform operator and $z$ is a complex 
variable.

In this work we harness three properties of the z-transform: 
linearity, time shifting, and convolution of sequences 
\citep{oppenheim1999discrete}. This converts the recursive nature of 
the derivations into simple algebraic operations, for example 
considering the reflex loop and assuming that $D$ and $A^{\prime}$ 
are zero, solving for the error signal yields:

\begin{align}\label{time-domain}
	\textit{time-domain} \quad \textit{\textbf{e[n]}} &=  i_{d}(t) - 
	r_{S}(r_{E}(r_{M}(e[n-1]))) \\
	 \textit{z-space} \quad \textit{\textbf{E(z)}} &=  I_{d}(z) -  
	 R_{S} R_{E} R_{M} \cdot z^{-1} E(z) = \frac{I_{d}}{1 + R_{S} 
	 R_{E} R_{M} z^{-1}} \label{z-space}
\end{align}

This shows that the error signal is function in time-domain, whereas, 
with z-transformation this translates into multiplication of transfer 
functions and sequences which allows for expression of $E(z)$ to be 
derived. In this work the $(z)$ symbol, representing the complex 
variable, is omitted for brevity.
\paragraph{The prime and modulate pathways:} As described in the 
introduction, only the sign of the error signal $E$ is used to 
generate the priming factor ($F_P$) for the \ac{BP}, while the global 
modulating factor ($F_{M}$) acts as a third factor where a filtered 
version of the
sensor inputs is used as a novelty or relevance signal which modulates
the learning with its amplitude. This is in line with the claim that
in particular dorsal striatal dopamine is more of a novelty or
salience detector \cite{Prescott2006} than an error signal, or
serotonin being a rectified version of the reward prediction error
\cite{Li2016}. These signals and their functionalities are described
in more details in the future sections.

\paragraph{The learning goal:} The aim of the learner is to produce a 
predictive signal $P$ such that the error signal $E$ is kept at zero 
persistently. In mathematical terms, this is analogous to minimising 
the absolute value of the error $|E|$; since this is a non 
differentiable function at $E=0$ the quadratic of $E$ is minimised 
instead:

\begin{align}\label{eq:goal}
\textit{Learning goal: } \quad P = \underset{p}{arg\,min} \textit{ } 
E^{2}
\end{align}

This is achieved by adjusting the internal parameters of the network 
that are introduced below.

\begin{figure}[H]
  \centering
  \includegraphics[width=1\textwidth]{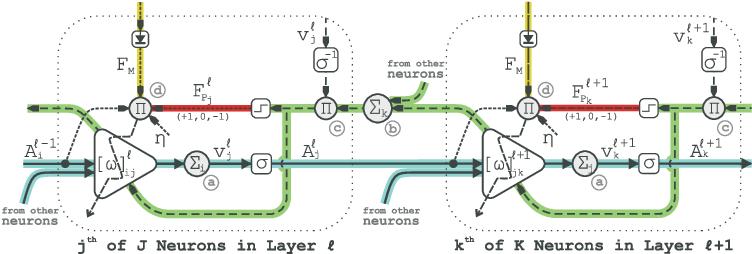}  
  \caption{\textit{Neuron connections in \ac{pam} network. Forward 
  propagation of inputs in shown with the left-to-right solid lines 
  highlighted in blue. $\sigma$ is the sigmoid activation function 
  and $A_{j}^{\ell}$ denotes the activation of $j^{th}$ neuron in 
  layer $\ell$. $[\omega]^{\ell}_{ij}$ is the weight matrix 
  associated with inputs $I$ inputs to this layer. The summation node 
  \textcircled{\small{a}} corresponds to 
  Equation~\ref{eq:forwardpass}. Backpropagation pathway is shown 
  with right-to-left dashed lines highlighted in green. The summation 
  at node \textcircled{\small{b}} and product at node  
  \textcircled{\small{c}} correspond to Equation~\ref{eq:backprop}. 
  The priming pathway is shown with short dashed lines highlighted in 
  red. This is the sign of the resulting value from the 
  backpropagation pathway, see Equation~\ref{eq:primingFactor}. The 
  modulating pathway is shown in long dashed lines that enter each 
  neuron from the environment, see 
  Equation~\ref{eq:modulatingFactor}. The priming and modulating 
  factors join at node \textcircled{\small{d}}, together with the 
  learning rate $\eta$ and the relevant input to the neuron 
  $A_{i}^{\ell -1}$, to drive the learning rule, corresponding to 
  Equation~\ref{eq:pamLearningrule}.  }}\label{fig:neuron}
\end{figure}

\paragraph{The neural network:} This paradigm employs a feed-forward 
neural network with fully connected layers. Figure~\ref{fig:neuron} 
shows the internal connections of two neurons in this network.

The forward propagation of activations $A$ is shown with solid 
right-arrows and is calculated as:\footnote{The superscripts denote 
the layer index, the first subscripts denote the neuron index, and 
the second subscripts denote the input (or the associated weight) 
index.}

\begin{align}\label{eq:forwardpass}
A_{j}^{\ell} &= \sigma(v_{j}^{\ell})  = 
\sigma(\Sigma_{i=1}^{I}\omega^{\ell}_{ij} \sigma(v_{i}^{\ell-1})) = 
\sigma(\Sigma_{i=1}^{I}\omega^{\ell}_{ij} A^{\ell-1}_{i})  \textit{ 
Where: } \ell = 1 \rightarrow L 
\end{align}

Where $\omega$ denotes the weights, $v$ is referred to as the sum 
output, and the $\sigma$ is the logistic function that maps the sum 
output of the neuron $v$ to the activation. The weighted sum takes 
place at summation points \textcircled{\small{a}}. Note that for 
$\ell = 1$, the predictive input information substitute for the input 
activations: $A^{0}_{i} = I'_{i}$ and, for $\ell = L$, the weighted 
sum of activations in the last layer yields the predictive signal in 
the closed-loop platform: $P = \Sigma_{x=0}^{X}(g_{x}A_{x}^{L})$, 
where $g_{x}$ is a weighting constant.

\section{Derivation of the learning rule}

The learning goal in Equation~\ref{eq:goal} is achieved through
adjustments of weights $\omega$ in Equation~\ref{eq:forwardpass}. At
each iteration, the error signal $E$ provides the network with
constructive feedback on the adequacy of the predictive signal $P$. 
Conventionally, the \ac{GDM} is
employed for weight optimisation; where the change to an arbitrary
weight is proportional to the sensitivity of $E^{2}$ with respect to
the sum output $v$ of the neuron containing that weight:

\begin{align}\label{eq:gdm}
\textit{Gradient descent method:} \quad \Delta \omega_{ij}^{\ell} 
\propto \frac{\partial E^{2}}{\partial v_{j}^{\ell}}
\end{align}

When differentiating in z-space we assume that the weight changes in 
time-domain are constant or significantly slower than the changes in 
the closed-loop system. (reference???). We seek an expression of this 
gradient in the context of closed-loop applications. With a similar 
approach to \cite{daryanavard2020closed}, this gradient is unravelled 
using the chain rule:

\begin{align}\label{eq:chainEPw}
\frac{\partial E^{2}}{\partial v_{j}^{\ell}} = 
\overbrace{\frac{\partial E^{2}}{\partial P}}^{Closed-loop} \cdot 
\overbrace{\frac{\partial P}{\partial v_{j}^{\ell}}}^{Network}
\end{align}

The former partial derivative solely relates to the dynamics of the 
closed-loop platform, whilst, the latter partial derivative relates 
to the inner connections of the network.

\paragraph{The closed-loop gradient ($\frac{\partial E^{2}}{\partial 
P}$):} Referring to Figure~\ref{fig:closedloop} (summation points 
\textcircled{\small{1}} and \textcircled{\small{2}}), the closed-loop 
expression of the error signal in z-space is derived as:

\begin{align}\label{eq:E}
\textit{\textbf{E}} &= I_{d} - I = I_{d} - 
R_{S}R_{E}(\overbrace{\textit{\textbf{E}}R_{M}}^{=A} + 
\overbrace{PL_{M}}^{=A'} + Dz^{-T}) \nonumber \\
&= \frac{I_{d} - R_{S}R_{E}(PL_{M} + Dz^{-T})}{1 + R_{S}R_{E}R_{M}}
\end{align}

Therefore, differentiation of quadratic of $E$ with respect to $P$ 
yields:

\begin{align}\label{eq:dEdP}
\textit{Closed-loop gradient: } \quad \frac{\partial E^{2}}{\partial 
P} = 2 E \frac{\partial E}{\partial P} = 2 E 
\frac{-R_{S}R_{E}L_{M}}{1 + R_{S}R_{E}R_{M}}
\end{align}

The value of the resulting fraction can be found experimentally by 
substituting $I_{d} = 0$, $D = 0$, and $P=1$ in Equation~\ref{eq:E} 
and measuring $E$; this is the closed loop gain of the system.

\paragraph{The network gradient ($\frac{\partial P}{\partial v}$):} 
From Equation~\ref{eq:forwardpass}, differentiating with respect to 
an arbitrary sum output $v^{\ell}_{j}$ results in a recursive 
expression:

\begin{align}\label{eq:backprop}
\textit{Network gradient:} \quad \frac{\partial P}{\partial 
v^{\ell}_{j}} = \sigma^{-1}(v^{\ell}_{j}) \cdot \Sigma_{k=0}^{K} 
(\omega^{\ell+1}_{jk} \frac{\partial P}{\partial v^{\ell+1}_{k}})
\end{align}

Where $\sigma^{-1}$ is the inverse logistic function. For the neurons 
in the final layer we have: $P = A^{L}_{x}$, therefore, this gradient 
is simply calculated as: $\sigma^{-1}(v^{L}_{x})$. This is fed 
through the network using the \acl{BP} technique. This chain of 
events is shown by dashed left-arrows in Figure~\ref{fig:neuron}, 
with the weighted sum and the product indicated at points 
\textcircled{\small{b}} and \textcircled{\small{c}}, respectively. 

With that, we have an expression of the gradient found in 
Equation~\ref{eq:gdm} which can be used for closed-loop learning 
applications. However, depending on the distribution of the weights 
and the topology of the neural network, the propagation can suffer 
from \acl{EVGP} \citep{pascanu2013difficulty, bengio1994learning, 
bengio1993problem}. In the following section we derive the \ac{pam} 
learning rule which is free from this issue.

\paragraph{\acl{pam} learning} In this work merely the \textit{sign} 
of the gradient $\frac{\partial E^{2}}{\partial v}$, found in 
Equation~\ref{eq:gdm}, is used for learning. This serves to 
'\textit{prime}' the weights to later undergo an increase, a 
decrease, or remain unchanged. Hence, it is referred to as the 
Priming Factor $F_{P}$ as seen in Figures~\ref{fig:closedloop} 
and~\ref{fig:neuron}:

\begin{align}\label{eq:primingFactor}
F_{P} = \delta(E) = \frac{\frac{\partial |E|^{2}}{\partial 
v}}{|\frac{\partial |E|^{2}}{\partial v}|} =
	\begin{cases}
      +1 & \text{\textit{primes $\omega$ to be increased}}\\
      0 & \text{\textit{primes $\omega$ to remain unchanged}}\\
      -1 & \text{\textit{primes $\omega$ to be decreased}}
    \end{cases} 
\end{align}

Once primed, the magnitude of weight change is dictated by a 
secondary signal that contains collective cues gathered from the 
environment informing the significance and/or relevance of the 
learning experience at any given instance of time. $R_{C}$ and 
$L_{C}$, shown in Figure~\ref{fig:closedloop}, are functions designed 
to extract relevant cues from the reflex and predictive loops, 
respectively. The correlation of these cues indicated at product 
point \textcircled{\small{4}}, modulates the magnitude of weight 
changes. Hence, this signal is referred to as the Modulating Factor 
$F_{M}$ (Figures~\ref{fig:closedloop} and~\ref{fig:neuron}):

\begin{align}\label{eq:modulatingFactor}
F_{M} = |ER_{C} \cdot I'L_{C}|
\end{align}

Finally, we redefine the weight change proportionality in 
Equation~\ref{eq:gdm}, and with the introduction of the learning rate 
$\eta$, we establish the update rule for this paradigm (refer to 
Equations~\ref{eq:forwardpass}, ~\ref{eq:chainEPw}, 
and~\ref{eq:dEdP}):

\begin{align}\label{eq:pamLearningrule}
\Delta \omega_{ij}^{\ell} &\propto \overbrace{\frac{2E \cdot 
\frac{-R_{S}R_{E}L_{M}}{1 + R_{S}R_{E}R_{M}} \cdot \frac{\partial 
P}{\partial v^{\ell}_{i}}}{|2E \cdot \frac{-R_{S}R_{E}L_{M}}{1 + 
R_{S}R_{E}R_{M}} \cdot \frac{\partial P}{\partial 
v^{\ell}_{i}}|}}^{\textit{priming factor}} \cdot \overbrace{|ER_{C} 
\cdot I'L_{C}|}^{\textit{modulating factor}} \nonumber \\ \Delta 
\omega_{ij}^{\ell} &\coloneqq \eta A^{\ell -1}_{i} {F_{P}}_{j}^{\ell} 
F_{M}
\end{align}

As explained above, ${F_{P}}_{j}^{\ell}$ is back-propagated through 
the layers, whilst, $F_{M}$ is available to all neurons in the 
network globally; Figure~\ref{fig:neuron} point 
\textcircled{\small{d}} illustrates the weight changes. This 
finalises the derivation of the learning rule for \acf{pam} paradigm.

\section{Experimentation platform: robotic navigation}

\begin{figure}[H]
  \centering
  \includegraphics[width=1\textwidth]{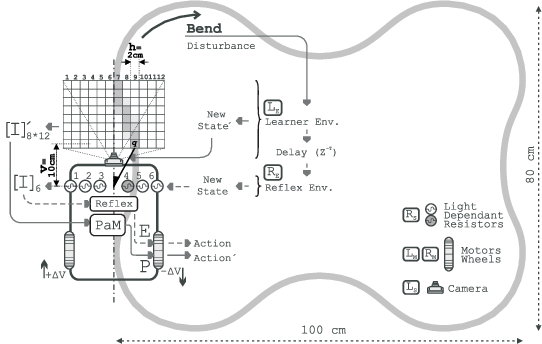}  
  \caption{\textit{Schematic of the experimental setup: A robot 
  placed on a canvas which navigates the path}}\label{fig:robot}
\end{figure}

Figure~\ref{fig:robot} illustrates the experimental setup where a 
robot is placed on a canvas with the task of following a path. The 
robot is fitted with a \ac{RPi} that hosts the learning algorithm as 
an external C++ library. The network is initialised with 10 hidden 
layers for the following set of experiments.

The chassis houses an array of \acfp{ldr}, a camera, two wheels with 
servo motors, and a battery bank to power the system. The equivalent 
closed-loop symbol of these components, as introduced in 
Figure~\ref{fig:closedloop}, are shown here.

\paragraph{\aclp{ldr}:} These sensors measure the \acp{gsv} $I_{i}$ 
of the surface underneath them and thus, monitor the position of the 
robot with respect to the path. The desired state is a symmetrical 
stand, therefore, a nonsymmetrical alignment results in the 
generation of a non-zero error signal as:

\begin{align}\label{eq:eTime}
E = a(I_{1}-I_{6}) + b(I_{2}-I_{5}) + c(I_{3}-I_{4}) \quad 
\textit{where: } a > b > c
\end{align}

The weighting registers the degree of deviation and enables sharp, 
medium, or gentle steering. 

\paragraph{Servo motors:} This error signal is used to recover the 
desired symmetrical state by altering the speed of motors:

\begin{align}\label{eq:motors}
V_{left} = V_{0} - \overbrace{d E}^{Reflex} - \overbrace{e 
P}^{Network} &= V_{0} - \Delta V \quad  \\ \nonumber \quad V_{right} 
&= V_{0} + \Delta V
\end{align}

Where $V_{0} = 7 \textit{\footnotesize{$[\frac{cm}{s}]$}}$ and $d$ is 
an experimental constant. This describes the functionality of the 
reflex loop in this application.

\paragraph{The camera} The view of the road ahead is captured by the 
camera in the form of a matrix of \acp{gsv} $[I]^{\prime}_{8*12}$ and 
is fed into the neural network.

In addition to the error signal, the predictive output of the network 
$P$ is used for steering as in Equation~\ref{eq:motors}, where $e$ is 
a scaling factor found experimentally. The error signal in 
Equation~\ref{eq:eTime} is used to train the network. Below, we 
derive the learning rule for this application.

\paragraph{The priming factor:} The closed loop gain derived in 
Equation~\ref{eq:dEdP} is found to be $\approx 0.97$ for this setup. 
Therefore, the Priming Factor for this learning is: $F_{P} = 1.94 E 
\frac{\partial P}{\partial v}$. Where the error signal and the 
partial derivative are calculated form Equations~\ref{eq:eTime} 
and~\ref{eq:backprop}, respectively.

\paragraph{The modulating factor:} When $E > 0$, a positive change 
indicates further deviation whilst a negative change implies 
recentering. In contrary, when $E < 0$, a negative change indicates 
further deviation from the path whilst a positive change implies 
recentering. Therefore, the product of the two, $E\frac{\partial 
E}{\partial t}$, signals the worsening or bettering of the 
performance when its value is positive or negative, respectively. The 
exponential of this product is the cue extracted from the reflex 
inputs $[I]_{6}$ signifying the learning at each iteration: $E \cdot 
R_{C} = e^{E\frac{\partial E}{\partial t}}$

Thus the learning is weakly modulated when the navigation is 
improving and is strongly modulated as the navigation worsens. The 
cue extracted from the predictive inputs $[I]'_{8*12}$ is the angle 
of deviation. The vertical distance of the bottom row of this matrix 
to the sensors' location $v$ and the horizontal spread $h$ of each 
pixel are measured. By having the index $q$ of the pixel that 
captures the path the angle of deviation is found as: $I^{\prime} 
\cdot L_{C} = \arctan{\frac{h \cdot |6.5-q|}{v}}$

The modulating factor is defined as the correlation of the 
aforementioned cues. Therefore, the update rule for this application 
is:

\begin{align}
\Delta \omega_{ij}^{\ell} &\coloneqq \eta A^{\ell -1}_{i} \cdot 1.94 
E \frac{\partial P}{\partial v_{j}^{\ell}} \cdot e^{E\frac{\partial 
E}{\partial t}} \cdot  \arctan{\frac{h \cdot |6.5-q|}{v}} 
\end{align}

\section{Results: comparison of \ac{GDM} and \ac{pam}}

The performance of \ac{pam} paradigm is compared to that of the 
closed-loop \ac{GDM} bench-marked in \cite{daryanavard2020closed}. 
Successful learning is defined as the state where the error's moving 
average over $10$ seconds, $0.1\int_{t-10}^{t}|E|dt $, falls below 
$1\%$ of its maximum value during the trial.

\begin{figure}[H]
  \centering
  \includegraphics[width=1\textwidth]{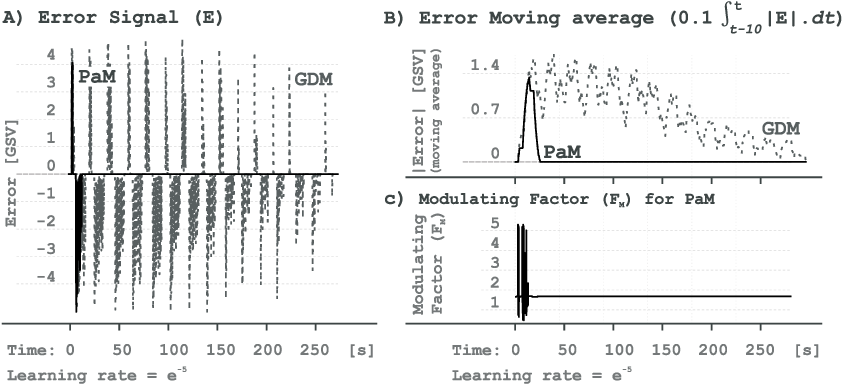}  
  \caption{\textit{A pair of comparative learning trials with 
  learning rate of $\eta = e^{-5}$}}\label{fig:errorSlow}
\end{figure}

The dashed traces in Figure~\ref{fig:errorSlow}A and B show the error 
signal $E$ and its moving average during a \ac{GDM} trial with the 
learning rate\footnote{This is the natural number $e = 2.71828$} of 
$\eta = e^{-5}$; the success condition is reached at time $t= 266.3 
[s]$. The black traces in these figures show the error signal $E$ and 
its moving average during a trial with \ac{pam}; the success 
condition is achieved at time $t= 23.1 [s]$. The modulating factor 
$F_{M}$ for this trial is shown in figure~\ref{fig:errorSlow}C. This 
pair of trials shows a significant improvement in the speed of 
learning and navigational performance of the robot.

\begin{figure}[H]
  \centering
  \includegraphics[width=1\textwidth]{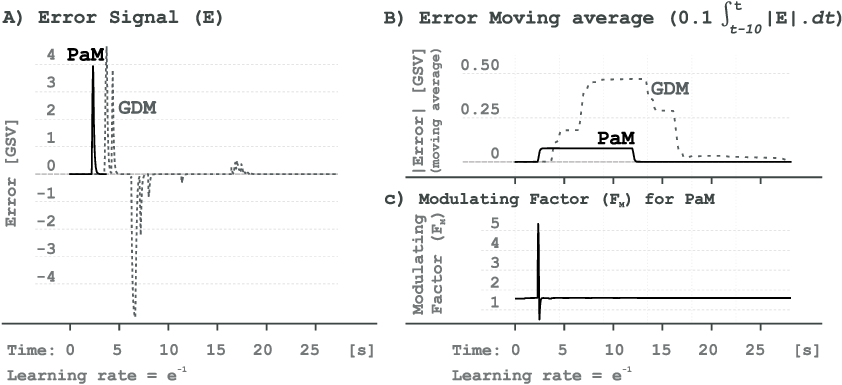}  
  \caption{\textit{A pair of comparative learning trials with 
  learning rate of $\eta = e^{-1}$}}\label{fig:fast}
\end{figure}

A second pair of trials with higher learning rate of $\eta = e^{-1}$ 
is shown in figure~\ref{fig:fast}; this shows a close to one-shot 
learning performance for \ac{pam}. Such comparative trial pairs were 
repeated $50$ times across learning rates of: $\eta = 
[e^{-5},e^{-4},e^{-3},e^{-2},e^{-1}]$.

\begin{figure}[H]
  \centering
  \includegraphics[width=1\textwidth]{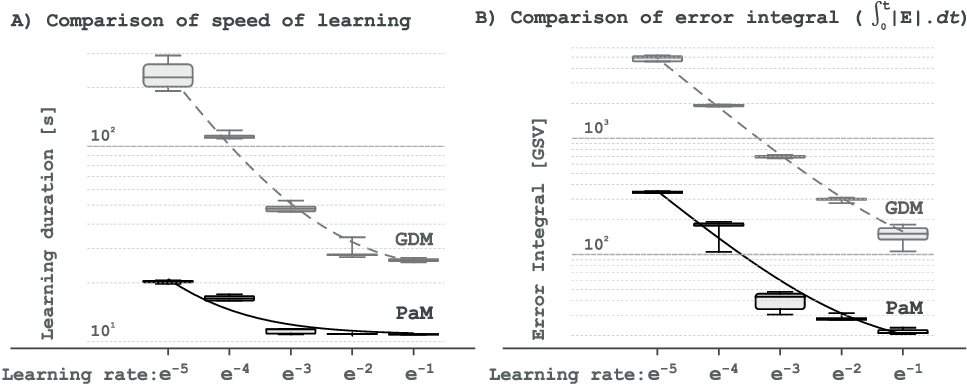}  
  \caption{\textit{comparison of learning speed and total error over 
  a range of 5 different learning rates}}\label{fig:results}
\end{figure}

Figure~\ref{fig:results}A shows the time taken for the robot to 
obtain the success state during the trials with \ac{pam} (black 
trace) compared to that of the \ac{GDM} (dashed trace). This shows 
consistency in the trend observed in Figures~\ref{fig:errorSlow} 
and~\ref{fig:fast}, in that, the \ac{pam} paradigm is significantly 
faster than its \ac{GDM} counterpart. Both methods demonstrate a 
faster learning with higher learning rates, however, the performance 
of \ac{GDM} is influenced by changes in the learning rate to a 
greater degree, given by the deflection of the fitted curve. 
Figure~\ref{fig:results}B show the corresponding error integrals, 
$\int_{0}^{t}|E|dt$, for these trials. As anticipated, the 
accumulation of the error signal is greater in trials with slower 
learning rate. Although the total accumulation of error is 
significantly smaller in \ac{pam} trials, in both methods this 
parameter is influenced by the learning rates to the same degree; 
inferred from the slope of the fitted curves.

\section{Discussion}

In this work we presented a novel closed loop algorithm which learns
the forward model of a reflex \cite{Porr2002nips}. These models play
an important role in robotic and biological motor control
\citep{wolpert1998multiple,wolpert2001perspectives,haruno2001mosaic,nakanishi2004feedback}
where they guarantee, for example, an optimal trajectory. Previous
work in this area used shallow networks \citep{kulvicius2007chained},
filter banks \cite{Porr2002nips} or single layers to perform
predictive control \citep{nakanishi2004feedback,maffei2017perceptual}
and it was not possible to employ deeper structures. On the other
hand, model free closed-loop learning has been using more complex
network structures such as deep learning in combination with
Q-learning \citep{guo2014,bansal2016learning}. Here, we demonstrate
that model based closed-loop learning can also benefit from deep
learning and thus a combination of both is more powerful
\citep{botvinick2019reinforcement}. However, given the fast learning 
of this model it can be prone to converging to a local minima. 
Whether or not deep learning is
biologically realistic has been debated for many years where the main 
issue
is the requirement of weight symmetry for forward- and backward-pass
which limits its plausibility to only a few layers
\cite{Lillicrap2016}. However, if the error is merely transmitted as a
sign, this weight symmetry can be relaxed as long as there are
interconnections between the top/down and bottom/up pathways
guaranteeing the correct sign of the learning
\cite{larkum2013cellular}. In the context of neuroscience, this means
that the bottom up pathway just controls long term potentiation (LTP)
or long term depression (LTD) while neuromodulators, in particular
serotonin as a rectified reward prediction error, can control the
speed of the learning \cite{Iigaya2018} as a third factor
\citep{Li2016}. Thus, in particular for cortical processing where
serotonin is more prominent than dopamine and deep neuronal structures
exist, a combination of local and global learning is a compelling fit
for neuroscience, in addition to its application in machine learning 
and robotic navigation.

\section*{Acknowledgements}
We would like to acknowledge Jarez Patel for his valuable 
intellectual and technical input for the making of the robotic 
platform.

\bibliographystyle{apa}
\bibliography{allTheRefs}

\end{document}